\documentclass[10pt,conference]{IEEEtran}
\IEEEoverridecommandlockouts
\usepackage{cite}
\usepackage{amsmath,amssymb,amsfonts}
\usepackage{algorithmic}
\usepackage{graphicx}
\usepackage{textcomp}
\usepackage{xcolor}
\usepackage{stfloats}
\usepackage{caption}   
\usepackage{booktabs}  
\usepackage{multirow} 
\usepackage[bookmarks=true,bookmarksnumbered=true,pdfstartview=FitH]{hyperref}
\hypersetup{colorlinks=true,linkcolor=blue,citecolor=blue,urlcolor=blue}
\def\BibTeX{{\rm B\kern-.05em{\sc i\kern-.025em b}\kern-.08em
    T\kern-.1667em\lower.7ex\hbox{E}\kern-.125emX}}
\newif\ifMemCamTeaserInserted
\MemCamTeaserInsertedfalse
\AtBeginDocument{%
  \let\MemCamOldtwocolumn\twocolumn
  \renewcommand\twocolumn[1][]{%
    \ifMemCamTeaserInserted
      \MemCamOldtwocolumn[{#1}]%
    \else
      \MemCamTeaserInsertedtrue
      \MemCamOldtwocolumn[{#1}{%
        \centering
        \includegraphics[width=\textwidth]{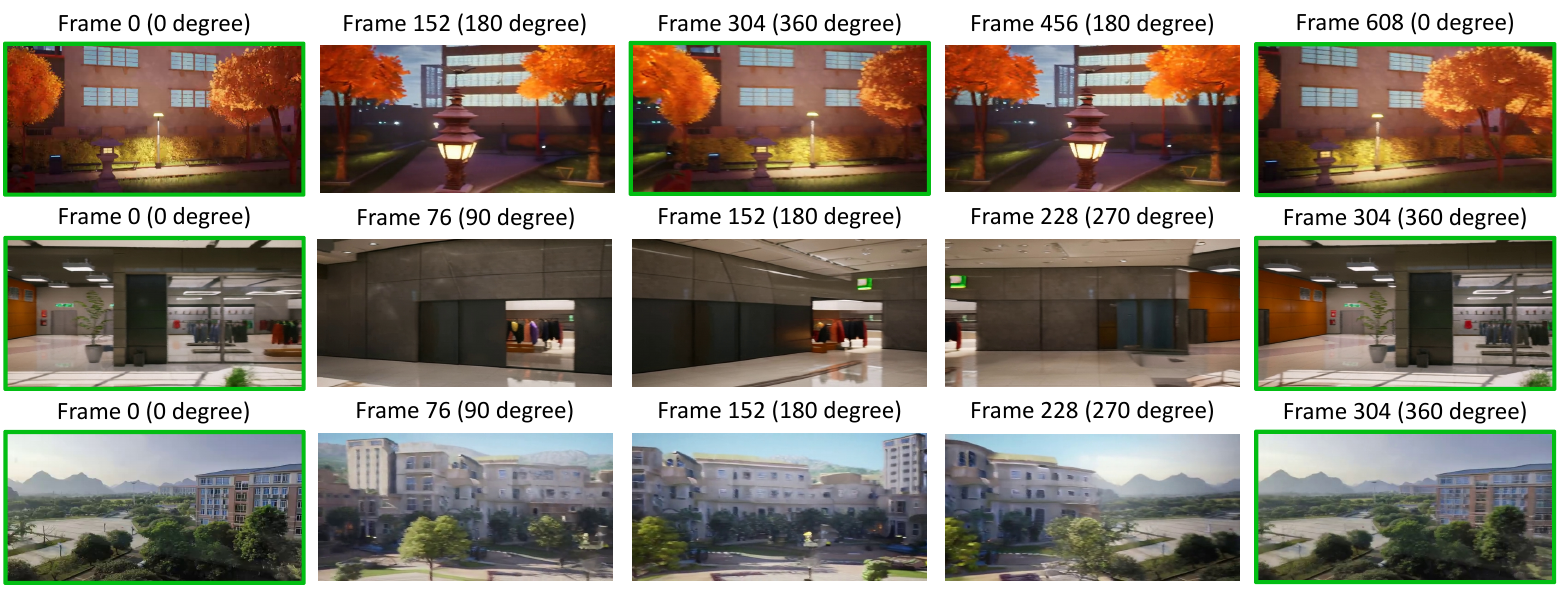}
        \captionof{figure}{\textbf{MemCam} generates videos with high scene consistency over long time horizons and under large camera rotations. Rows 1-2 show results on the test split of our dataset, including 360° round-trip and 360° single-direction rotation. Row 3 presents open-domain real-world scenes, demonstrating that MemCam maintains long-term scene consistency across varied scenes.}
        \label{fig:teaser}
        \vspace{1em}
        }]%
    \fi
  }%
}
\begin{document}

\title{MemCam: Memory-Augmented Camera Control for Consistent Video Generation}
\author{\IEEEauthorblockN{Xinhang Gao, Junlin Guan, Shuhan Luo, Wenzhuo Li, Guanghuan Tan, and Jiacheng Wang}\IEEEauthorblockA{
Guilin University of Electronic Technology, China\\
Corresponding author: Junlin Guan (e-mail: guanjunlin@guet.edu.cn)}}
\maketitle

\begin{abstract}
Interactive video generation has significant potential for scene simulation and video creation. However, existing methods often struggle with maintaining scene consistency during long video generation under dynamic camera control due to limited contextual information. To address this challenge, we propose MemCam, a memory-augmented interactive video generation approach that treats previously generated frames as external memory and leverages them as contextual conditioning to achieve controllable camera viewpoints with high scene consistency. To enable longer and more relevant context, we design a context compression module that encodes memory frames into compact representations and employs co-visibility-based selection to dynamically retrieve the most relevant historical frames, thereby reducing computational overhead while enriching contextual information. Experiments on interactive video generation tasks show that MemCam significantly outperforms existing baseline methods as well as open-source state-of-the-art approaches in terms of scene consistency, particularly in long video scenarios with large camera rotations.
\end{abstract}

\begin{IEEEkeywords}
interactive video generation, camera control, diffusion model
\end{IEEEkeywords}

\section{Introduction}
Recent advances in video generation models \cite{wan2025wan,kong2024hunyuanvideo,yang2024compositional,bao2024vidu} have made substantial progress, enabling higher visual quality and realism in video synthesis. Within the broader landscape of video generation, the sub-field of interactive video generation has gained prominence as a key research focus. Its growing importance stems from promising applications in areas such as video or game scene generation \cite{yu2025gamefactory,valevski2024diffusion,li2025vmem} and world simulation \cite{hu2023gaia1,russell2025gaia2,xiao2025worldmem}. Recent research on long video generation \cite{gao2025longvie,gao2025longvie2,yang2025longlive,zhang2025packing} has provided valuable methodologies to maintain coherence over extended sequences, thus significantly supporting and accelerating developments in interactive video generation.

Despite these advances, existing methods still face significant challenges when generating long interactive videos: they struggle to maintain the consistency of the scene content over extended temporal sequences \cite{decart2024oasis,song2025history}. This limitation manifests itself as the model's tendency to forget previously generated content during continuous synthesis. When the camera viewpoint returns to a previously displayed region after complex motion, it often leads to content discrepancies within the same scene at different time points \cite{yu2025gamefactory,valevski2024diffusion,decart2024oasis}. This is because these methods lack memory capability—they can only rely on camera information without explicit memory of past content, such as CameraCtrl \cite{he2024cameractrl}, or depend on a fixed-length context window of limited scope, as in DFoT \cite{song2025history}. Some approaches attempt to improve this by incorporating 3D reconstruction, like GeometryForcing \cite{wu2025geometryforcing}, but still face issues such as error accumulation and inherent performance limitations of the reconstruction models themselves.

\begin{figure*}[!t]
    \centering
    \includegraphics[width=\textwidth]{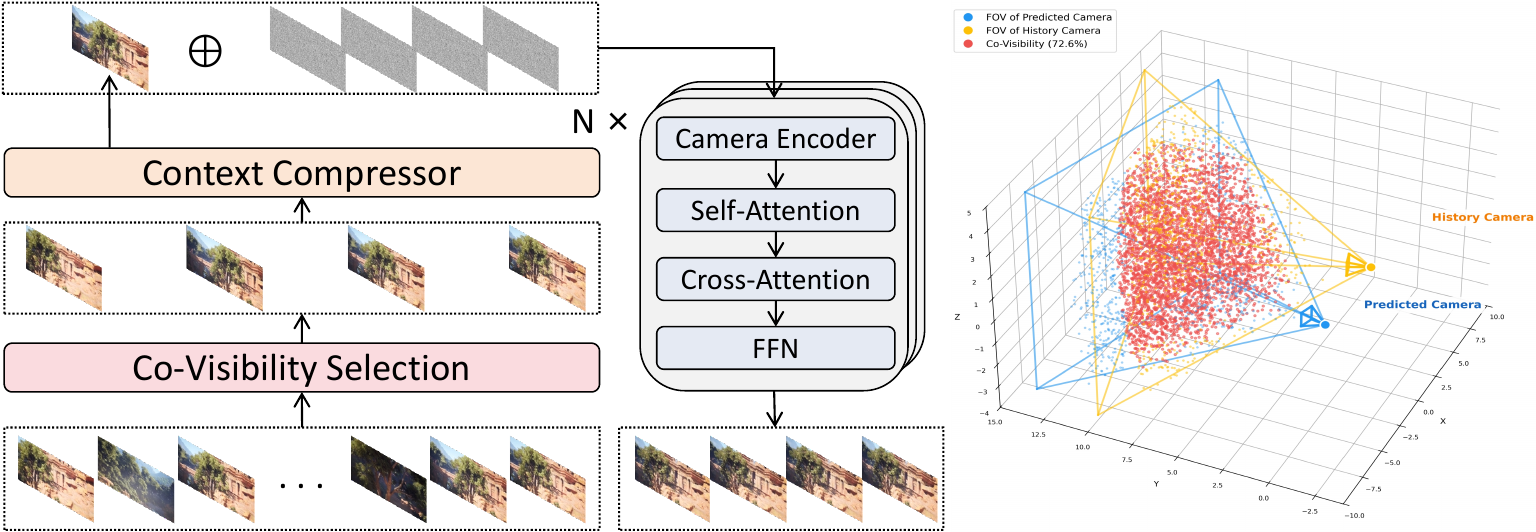}
    \caption{\textbf{Methodology.} (Left) Overview of MemCam: the Context Compressor encodes historical frames selected via co-visibility into compact representations, which are concatenated with the noisy prediction sequence and fed into the DiT Block. (Right) Illustration of co-visibility computation between predicted and historical camera FOVs.}
    \label{fig:methodology}
\end{figure*}

In this paper, we propose MemCam, a memory-augmented framework for scene-consistent interactive video generation. MemCam explicitly maintains historical frames as contextual memory and leverages them to condition future predictions, enabling long-term consistency under extended temporal horizons and large camera rotations. Without requiring additional 3D reconstruction modules, MemCam introduces a context compression module that effectively compresses historical frames selected via co-visibility, preserving scene structure across time. The overall framework is illustrated in Fig.~\ref{fig:methodology}. Our implementation is publicly available at \url{https://github.com/newhorizon2005/MemCam}.

Our main contributions can be summarized as follows:
\begin{itemize}
    \item We propose MemCam, a memory-augmented framework that leverages historical frames and positional information as contextual cues to enable scene-consistent interactive video generation.
    \item We design a context compression module that, through a co-visibility-based context selection strategy, effectively filters and compresses historical frame information to provide rich and diverse contextual support for long-term video prediction.
    \item Experiments demonstrate that our method excels in long interactive video generation, significantly outperforming baseline approaches.
\end{itemize}

\section{Related Work}
\subsection{Video Generation Models}
Video generation models are developing rapidly, with mainstream model architectures largely based on diffusion models \cite{ho2020denoising,lipman2022flow,liu2022flow}. Early diffusion-based video generation methods primarily employed U-Net-based \cite{ronneberger2015unet} architectures to extend image diffusion models, where temporal information was modeled using short-frame windows or temporal convolutions \cite{ho2022video, singer2022makeavideo}. Building on this progress, Transformer-based architectures, particularly Diffusion Transformers (DiT) \cite{peebles2023dit}, as adopted by recent models \cite{wan2025wan,kong2024hunyuanvideo,yang2025cogvideo}, replaced convolutional backbones with global self-attention mechanisms, allowing for more powerful spatiotemporal modeling and higher-quality video synthesis. Meanwhile, other architectural designs have also been explored, including hybrid attention mechanisms in video diffusion \cite{wang2024swapattention} and decomposed diffusion models \cite{luo2023videofusion}. These advances continue to improve the generative quality, temporal consistency, and controllability of video generation models.

\subsection{Interactive Video Generation}
The field of interactive video generation, where users provide control signals to guide the creation process, is also rapidly evolving. Existing models integrate various signals to enable fine-grained control over video content, such as camera trajectories \cite{he2024cameractrl,bai2025recammaster} and object motion \cite{wang2024motionctrl,feng2024matrix}. Recent work has explored the use of context guidance for interactive video generation. DFoT \cite{song2025history} introduces a training paradigm called "Diffusion Forcing," which independently perturbs the noise levels of different frames, allowing the model to extract conditional information from historical frames for flexible long-term video generation. GeometryForcing \cite{wu2025geometryforcing} takes a different approach by incorporating explicit 3D reconstruction as geometric constraints to enforce multi-view consistency during generation. Context-as-Memory \cite{yu2025cam} achieves scene-consistent interactive video generation without explicit 3D representations, but its code and model weights have not been publicly released.

\section{Methodology}
\subsection{Context Compression Module}
Directly conditioning on all historical frames is not only computationally inefficient, but may also introduce negative effects due to redundant or irrelevant information. To address this issue, we design a context compression module to aggregate contextual content. Specifically, we train a convolutional neural network to process the context portion, replacing the patchify layer of the base model. The weights of this network are initialized by extending the pre-trained model's patchify layer, ensuring meaningful outputs from the early stages of training. Given the lack of temporal correlation among historical frames, we keep the temporal dimension uncompressed, while the compression ratios for the spatial length and width dimensions are set to 2. The processed context tokens are only one-fourth the length of the unprocessed ones. For positional encoding, we use Rotary Positional Encoding (RoPE) \cite{su2023rope}, which allows variable input lengths. We keep the positional encoding for the prediction sequence consistent with the pre-training setup, assign new positional encodings to the context, and down-sample the spatial positional encodings of the context via pooling to align with the prediction sequence. Context compression enables the model to access more contextual frames simultaneously, thereby enriching the diversity of contextual information while reducing computational cost.

\subsection{Camera Control}
Camera control capability serves as the foundation for interactive video generation models. Following the work of RecamMaster \cite{bai2025recammaster}, we add a single-layer MLP called Camera Encoder to each DiT Block. For any input sequence, we obtain camera information $\text{cam} \in \mathbb{R}^{F \times 12}$, which is derived by flattening a $3 \times 4$ matrix composed of $\mathbf{R} | \mathbf{t}$, where $\mathbf{R} \in \mathbb{R}^{3 \times 3}$ is the rotation matrix, $\mathbf{t} \in \mathbb{R}^{3 \times 1}$ is the translation vector, and $\mathbf{R} | \mathbf{t}$ denotes their concatenation. This camera information is mapped through the encoder to the channel dimension of the model's main feature, expanded via repetition, and then added element-wise to the main feature. The operation is formulated as:

\begin{equation}
\mathbf{F}_{\text{out}} = \mathbf{F}_{\text{in}} + \text{CameraEncoder}(\text{cam}),
\end{equation}

where $\mathbf{F}_{\text{in}}$ denotes the input to the DiT Block, and $\mathbf{F}_{\text{out}}$ represents the features input to the self-attention layer after camera conditioning is incorporated.

\subsection{Memory Module with Context Selection}  
We introduce a memory module that explicitly incorporates historical frames as contextual guidance into the generation process. Specifically, we maintain a sequence of historical frames along with their corresponding camera information, called the historical sequence. For each frame in the predicted sequence, we leverage its camera information to compute the co-visibility score with every frame in the historical sequence, measuring the overlap between their respective fields of view. Based on this score, we select a subset of historical frames for each predicted frame and concatenate them as contextual input for the model.

We estimate the co-visibility between two camera poses via Monte Carlo sampling of 3D points to approximate the field-of-view (FOV) overlap. Given camera positions and orientations, the co-visibility is defined as the Intersection over Union (IoU) of the visible point sets:

\begin{equation}
\text{IoU}(\mathcal{C}_1, \mathcal{C}_2) = \frac{\sum_{i=1}^N \mathcal{V}_1(\mathbf{x}_i) \land \mathcal{V}_2(\mathbf{x}_i)}{\sum_{i=1}^N \mathcal{V}_1(\mathbf{x}_i) \lor \mathcal{V}_2(\mathbf{x}_i)} \in [0, 1],
\end{equation}

where $\mathcal{C}_1$ and $\mathcal{C}_2$ denote the two camera configurations, $N$ is the number of sampled points $\mathbf{x}_i$, and $\mathcal{V}_j(\mathbf{x}_i)$ indicates the visibility of point $\mathbf{x}_i$ from camera $j$. We set $N=10^4$ to achieve a robust approximation.

\subsection{Training and Inference}
During training, we randomly sample a clip from a scene as the target sequence. The first frame of the target sequence serves as a fixed context frame to ensure smooth transitions between consecutive video segments. This fixed context frame is concatenated directly with the prediction sequence and jointly encoded by the 3D VAE \cite{kingma2013vae}, without passing through the compression module. For each frame in the prediction sequence, we assign one context frame selected from the remaining frames in the scene based on co-visibility: any frame with non-zero overlap is eligible for selection. These selected context frames are individually encoded by the 3D VAE and then processed through our context compression module. To support image-to-video generation, with 10\% probability, we zero-pad all the selected context to simulate scenarios where no historical frames are available.

During inference, we adopt a segment-wise generation approach. Unlike training, inference selects the context frame with the highest co-visibility score for each predicted frame from historical frames, ensuring the most relevant content is retrieved. The memory is updated after each segment is generated, and this process iterates until the full video is completed.

\begin{figure*}[!t]
    \centering
    \includegraphics[width=\textwidth]{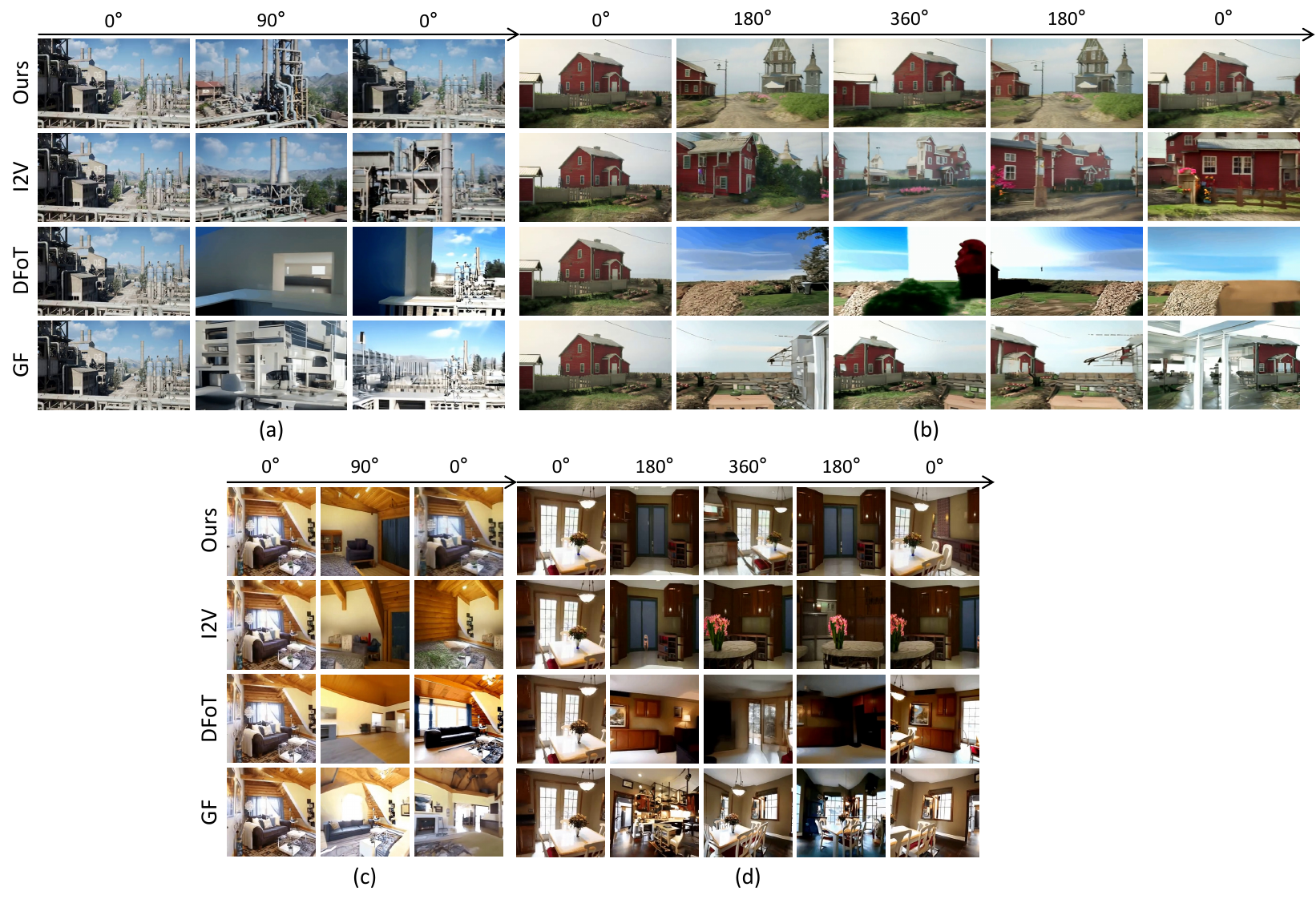}
    \caption{\textbf{Qualitative Comparison Results.} (a) and (b) are evaluated on the Context-as-Memory dataset, and (c) and (d) on RealEstate10K. MemCam achieves superior performance in scene memory retention and overall generation quality. In contrast, other methods exhibit varying degrees of scene inconsistency due to insufficient utilization of contextual information.}
    \label{fig:qualitative}
\end{figure*}

\section{Experiment}
\subsection{Experiment Settings}
\noindent\textbf{Implementation Details.} Our method is built on the Wan2.1 1.3B Text-to-Video Diffusion Transformer \cite{wan2025wan}. We train on the Context-as-Memory dataset \cite{yu2025cam}, containing 100 scenes with 7901 frames each. Each training instance consists of 77 frames at $640\times352$ resolution: the first frame serves as fixed context, and the remaining 76 frames form the prediction sequence. For each predicted frame, one context frame is selected from the other frames in the scene based on co-visibility. The fixed context and prediction sequence are jointly encoded by the 3D VAE into 20 latent frames, while the 76 selected context frames are individually encoded and processed through our compression module. All model parameters are trainable. The model was trained for 20,000 iterations using AdamW \cite{loshchilov2019adamw} with learning rate $1 \times 10^{-5}$, batch size 16, on 2 NVIDIA H20 GPUs. During inference, we use 50 denoising steps. We evaluate on 5\% of Context-as-Memory scenes as test set, and additionally on RealEstate10K \cite{zhou2018stereo} for zero-shot generalization.

\noindent\textbf{Evaluation Metrics.} To effectively evaluate the memory capacity of the video generation model, we adopt the following two assessment methods: (1) 90-degree round-trip generation: A video of $1 + 76 \times 2 = 153$ frames is generated, where the camera trajectory first rotates 90 degrees to the right and then rotates 90 degrees to the left back to the origin. (2) 360-degree round-trip generation: A video of $1 + 76 \times 8 = 609$ frames is generated, where the camera first completes a full 360-degree clockwise rotation and then a full 360-degree counterclockwise rotation, both returning to the starting viewpoint. These two methods effectively test the memory capacity of the model. Each generated video can be viewed as two sequences. Ideally, if one sequence is temporally reversed, it should be identical to the other sequence. We quantify this difference by using the PSNR, SSIM \cite{2004SSIM}, and LPIPS \cite{zhang2018LPIPS} metrics and employ FVD \cite{unterthiner2019FVD} to assess overall video quality.

\noindent\textbf{Comparison Methods.} We compare MemCam with the following methods: (1) Using only the first frame as context (I2V): implemented on our base model with the same training setup as MemCam; (2) Diffusion Forcing Transformer (DFoT) \cite{song2025history}: based on a fixed-length context window; (3) GeometryForcing (GF) \cite{wu2025geometryforcing}: based on 3D representation reconstruction.

\begin{table*}[!t]
    \centering
    \caption{\textbf{Quantitative Comparison Results.} We evaluate on two datasets with 90° and 360° round-trip benchmarks. $\uparrow$ means higher is better, $\downarrow$ means lower is better. Best results are in \textbf{bold}, second best are \underline{underlined}.}
    \label{tab:quantitative}
    \begin{tabular}{l|cccc|cccc}
        \toprule
        \multicolumn{9}{c}{\textbf{Context-as-Memory Dataset}} \\
        \midrule
        \multirow{2}{*}{Methods} & \multicolumn{4}{c|}{90° Round-trip} & \multicolumn{4}{c}{360° Round-trip} \\
        & PSNR$\uparrow$ & SSIM$\uparrow$ & LPIPS$\downarrow$ & FVD$\downarrow$ 
        & PSNR$\uparrow$ & SSIM$\uparrow$ & LPIPS$\downarrow$ & FVD$\downarrow$ \\
        \midrule
        I2V & 15.81 & 0.452 & 0.470 & 528.51 & 9.75 & 0.332 & 0.603 & 988.82 \\
        DFoT & \underline{16.76} & 0.474 & 0.393 & 683.59 & 8.94 & 0.252 & 0.613 & 1188.34 \\
        GF & 16.57 & \underline{0.486} & \textbf{0.348} & \underline{557.66} & \underline{10.07} & \underline{0.402} & \underline{0.565} & \underline{852.05} \\
        \textbf{MemCam (Ours)} & \textbf{17.83} & \textbf{0.506} & \underline{0.357} & \textbf{215.71} & \textbf{14.81} & \textbf{0.423} & \textbf{0.504} & \textbf{167.87} \\
        \midrule
        \multicolumn{9}{c}{\textbf{RealEstate10K Dataset}} \\
        \midrule
        \multirow{2}{*}{Methods} & \multicolumn{4}{c|}{90° Round-trip} & \multicolumn{4}{c}{360° Round-trip} \\
        & PSNR$\uparrow$ & SSIM$\uparrow$ & LPIPS$\downarrow$ & FVD$\downarrow$ 
        & PSNR$\uparrow$ & SSIM$\uparrow$ & LPIPS$\downarrow$ & FVD$\downarrow$ \\
        \midrule
        I2V & 16.26 & 0.488 & 0.362 & 552.01 & 10.16 & 0.284 & 0.611 & 789.62 \\
        DFoT & 17.17 & 0.505 & 0.399 & 539.89 & 10.43 & 0.281 & 0.564 & 1002.39 \\
        GF & \textbf{17.70} & \textbf{0.597} & \underline{0.316} & \underline{519.78} & \underline{11.19} & \underline{0.379} & \underline{0.405} & \underline{419.60} \\
        \textbf{MemCam (Ours)} & \underline{17.61} & \underline{0.544} & \textbf{0.314} & \textbf{269.82} & \textbf{16.52} & \textbf{0.550} & \textbf{0.400} & \textbf{131.96} \\
        \bottomrule
    \end{tabular}
\end{table*}

\subsection{Main Results}
\noindent\textbf{Quantitative Results.} As shown in Table~\ref{tab:quantitative}, MemCam achieves competitive or superior performance across most metrics, with particularly significant gains in the 360° scenario where longer duration and larger camera rotation pose greater challenges. The I2V baseline, lacking historical memory, generally performs poorly. DFoT, constrained by its limited context window, fails to retain early content in long sequences. GF achieves competitive PSNR/SSIM on 90° tasks, but its performance drops notably in 360° scenarios as reconstruction errors accumulate. MemCam achieves the best FVD across all settings, indicating superior temporal consistency and visual quality. The results show that effective utilization of historical frames through compression and retrieval is crucial for memory retention and generation quality.

\noindent\textbf{Qualitative Results.} As shown in Fig.~\ref{fig:qualitative}, MemCam faithfully preserves the scene structure even after large camera rotations, while other methods exhibit noticeable drift and content hallucination. When the camera returns to the starting viewpoint, MemCam reconstructs the original scene appearance, whereas baselines produce inconsistent textures or altered layouts. Sufficient context conditioning not only strengthens memory retention, but also suppresses error accumulation in long-range generation.

\subsection{Ablation Study}
All ablation experiments are conducted on the Context-as-Memory dataset.

\noindent\textbf{Context Selection Strategy.} We compare different strategies for selecting context frames. ``Recent'' selects the most recent historical frames; ``Random'' randomly samples from historical frames; ``TopK'' aggregates all frames that overlap with any frame in the prediction sequence, ranks them by total occurrence count, and selects the top-ranked ones. As shown in Table~\ref{tab:ablation_selection}, our selection strategy consistently outperforms all alternatives. On the 90° benchmark, Recent performs comparably to Ours since the limited camera rotation ensures that recent frames remain spatially relevant. On the 360° benchmark, Recent degrades dramatically as it cannot retrieve content from distant viewpoints. Random, despite lacking targeted selection, outperforms Recent because it can access frames from earlier parts of the video that cover distant viewpoints. TopK tends to favor frames from the middle of the sequence, resulting in uneven coverage. Our method achieves the best results by providing both relevance and uniform coverage through per-frame dynamic selection.

\begin{table}[t]
    \centering
    \caption{\textbf{Ablation on Context Selection Strategy.}}
    \label{tab:ablation_selection}
    \small
    \begin{tabular}{l|cccc}
        \toprule
        \multicolumn{5}{c}{\textbf{90° Round-trip}} \\
        \midrule
        Methods & PSNR$\uparrow$ & SSIM$\uparrow$ & LPIPS$\downarrow$ & FVD$\downarrow$ \\
        \midrule
        Recent & \underline{17.80} & 0.498 & \underline{0.375} & \underline{231.75} \\
        Random & 17.36 & \underline{0.502} & 0.400 & 237.87 \\
        TopK & 17.39 & 0.497 & 0.394 & 256.88 \\
        \textbf{Ours} & \textbf{17.83} & \textbf{0.506} & \textbf{0.357} & \textbf{215.71} \\
        \midrule
        \multicolumn{5}{c}{\textbf{360° Round-trip}} \\
        \midrule
        Methods & PSNR$\uparrow$ & SSIM$\uparrow$ & LPIPS$\downarrow$ & FVD$\downarrow$ \\
        \midrule
        Recent & 10.26 & 0.340 & 0.581 & 915.41 \\
        Random & 12.13 & \underline{0.417} & 0.511 & \underline{238.98} \\
        TopK & \underline{14.67} & 0.414 & \underline{0.506} & 262.52 \\
        \textbf{Ours} & \textbf{14.81} & \textbf{0.423} & \textbf{0.504} & \textbf{167.87} \\
        \bottomrule
    \end{tabular}
\end{table}

\noindent\textbf{Context Compression Module.} We evaluate the context compression module on the 360° round-trip benchmark by varying total context length (76, 38, or 19 frames, i.e., one context frame per 1, 2, or 4 predicted frames). ``None'' denotes direct concatenation without compression, while ``Ours'' applies our compression module. We report inference time as seconds per frame (s/frame). As shown in Table~\ref{tab:ablation_compression}, without compression, increasing context length yields marginal quality gains but incurs substantial computational overhead. With compression, longer context provides richer information that better tolerates the compact encoding, leading to improved performance. Notably, Ours-76 attains similar quality to None-76 while being nearly 5$\times$ faster, and even outperforms None-19 at comparable time cost. These results demonstrate that context compression effectively enables richer context without prohibitive computational burden.

\begin{table}[t]
    \centering
    \caption{\textbf{Ablation on Context Compression Module.}}
    \label{tab:ablation_compression}
    \small
    \begin{tabular}{l|ccccc}
        \toprule
        Methods & PSNR$\uparrow$ & SSIM$\uparrow$ & LPIPS$\downarrow$ & FVD$\downarrow$ & s/frame$\downarrow$ \\
        \midrule
        None-19 & 14.69 & 0.417 & 0.513 & 184.42 & 4.45 \\
        None-38 & 14.76 & 0.424 & 0.494 & 179.29 & 8.82 \\
        None-76 & 14.88 & 0.430 & 0.489 & 164.37 & 22.15 \\
        \midrule
        Ours-19 & 13.88 & 0.398 & 0.534 & 203.30 & 2.14 \\
        Ours-38 & 14.61 & 0.407 & 0.527 & 182.99 & 2.81 \\
        Ours-76 & 14.81 & 0.423 & 0.504 & 167.87 & 4.47 \\
        \bottomrule
    \end{tabular}
\end{table}

\section{Conclusion}
In this work, we propose MemCam, a memory-augmented framework for scene-consistent interactive video generation that treats previously generated frames as external memory. We design a context compression module that encodes historical frames into compact representations, enabling the model to incorporate richer contextual information while significantly reducing computational cost. Combined with a co-visibility-based context selection strategy, our framework effectively retrieves the most relevant historical frames for each predicted frame, enabling faithful scene reconstruction even under large camera rotations. Experiments on two datasets demonstrate that MemCam significantly outperforms existing methods, particularly in long video scenarios with large viewpoint changes.

\noindent\textbf{Limitations and Future Work.} The current inference speed is relatively slow due to the computational overhead of bidirectional attention. In future work, we plan to explore diffusion distillation for acceleration and scale up training with larger datasets to further improve generation quality.

\section*{Acknowledgment}
This work was supported by the National Innovation Training Program for College Students of Guilin University of Electronic Technology (Grant No. 202510595051).

\bibliographystyle{IEEEtran}
\bibliography{references}
\vspace{12pt}
\end{document}